\documentclass[letterpaper]{article}
\usepackage{iccc}

\usepackage[utf8]{inputenc}

\usepackage{times}
\usepackage{helvet}
\usepackage{courier}
\title{Automatic Dialect Adaptation in Finnish and its Effect on Perceived Creativity}
\author{Mika Hämäläinen\textsuperscript{1}\\
mika.hamalainen@helsinki.fi\\
\And
Niko Partanen\textsuperscript{2}\\
niko.partanen@helsinki.fi \\
\And
Khalid Alnajjar\textsuperscript{1}\\
khalid.alnajjar@helsinki.fi \\
\AND
Jack Rueter\textsuperscript{2}\\
jack.rueter@helsinki.fi \\
\And
Thierry Poibeau\textsuperscript{3}\\
thierry.poibeau@ens.fr \\
\AND
\vspace*{-6mm}\mbox{}\\
\vspace*{-0mm}\textsuperscript{1}Rootroo Ltd \& University of Helsinki,
\textsuperscript{2}University of Helsinki,\\ 
\textsuperscript{3}Lab. LATTICE, ENS/PSL \& CNRS \& Univ. Sorbonne nouvelle, FR
}
\begin{document} 
\maketitle
\begin{abstract}
\begin{quote}
We present a novel approach for adapting text written in standard Finnish to different dialects. We experiment with character level NMT models both by using a multi-dialectal and transfer learning approaches. The models are tested with over 20 different dialects. The results seem to favor transfer learning, although not strongly over the multi-dialectal approach. We study the influence dialectal adaptation has on perceived creativity of computer generated poetry. Our results suggest that the more the dialect deviates from the standard Finnish, the lower scores people tend to give on an existing evaluation metric. However, on a word association test, people associate \textit{creativity} and \textit{originality} more with dialect and \textit{fluency} more with standard Finnish.
\end{quote}
\end{abstract}

\section{Introduction}

We present a novel method for adapting text written in standard Finnish to different Finnish dialects. The models developed in this paper have been released in an open-source Python library\footnote{https://github.com/mikahama/murre} to boost the limited Finnish NLP resources, and to encourage both replication of the current study and further research in this topic. In addition to the new methodological contribution, we use our models to test the effect they have on perceived creativity of poems authored by a computationally creative system.

Finnish language exhibits numerous differences between colloquial spoken regional varieties and the written standard. This situation is a result of a long historical development. Literary Finnish variety known as Modern Finnish developed into its current form in late 19th century, after which the changes have been mainly in the details \cite[16]{hakkinen1994agricolasta}. Many of the changes have been lexical due to technical innovations and modernization of the society: orthographic spelling conventions have largely remained the same. Spoken Finnish, on the other hand, traditionally represents an areally divided dialect continuum, with several sharp boundaries, and many regions of gradual differentiation from one municipality to another municipality. 

Especially in the later parts of 21th century the spoken varieties have been leveling away from very specific local dialects, and although regional varieties still exist, most of the local varieties have certainly became endangered. Similar processes of dialect convergence have been reported from different regions in Europe, although with substantial variation \cite{auer2018dialect}. In the case of Finnish this has not, however, resulted in merging of the written and spoken standards, but the spoken Finnish has remained, to our day, very distinct from the written standard. In a late 1950s, a program was set up to document extant spoken dialects, with the goal of recording 30 hours of speech from each municipality. This work resulted in very large collections of dialectal recordings \cite[448-449]{lyytikainen1984suomen}. Many of these have been published, and some portion has also been manually normalized. Dataset used is described in more detail in Section Data and Preprocessing.

Finnish orthography is largely phonemic within the language variety used in that representation, although, as discussed above, the relationship to actual spoken Finnish is complicated. Phonemicity of the orthography is still a very important factor here, as the differences between different varieties are mainly displaying historically developed differences, and not orthographic particularities that would be essentially random from contemporary point of view. Thereby the differences between Finnish dialects, spoken Finnish and Standard Finnish are highly systematic and based to historical sound correspondences and sound changes, instead of more random adaptation of historical spelling conventions that would be typical for many languages.

Due to the phonemicity of the Finnish writing system, dialectal differences are also reflected in informal writing. People speaking a dialect oftentimes also write it as they would speak it when communicating with friends and family members. This is different from English in that, for example, although Australians and Americans pronounce the word \textit{today} differently, they would still write the word in the same way. In Finnish, such a dialectal difference would result in a different written form as well.

We hyphotesize that dialect increases the perceived value of computationally created artefacts. Dialectal text is something that people are not expecting from a machine as much as they would expect standard Finnish. The effect dialect has on results can be revealing of the shortcomings of evaluation methods used in the field.

\section{Related Work}

Text adaptation has received some research attention in the past. The task consists of adapting or transferring a text to a new form that follows a certain style or domain. As the particular task of dialect adaptation has not received a wide research interest, we dedicate this section in describing different text adaptation systems in a mode broad sense.

Adaptation of written language to a more spoken language style has previously been tackled as a lexical adaptation problem \cite{kaji2005lexical}. They use style and topic classification to gather data representing written and spoken language styles, thereafter, they learn the probabilities of lexemes occurring in both categories. This way they can learn the differences between the spoken and the written on a lexical level and use this information for style adaptation. The difference to our approach is that we approach the problem on a character level rather than lexical level. This makes it possible for our approach to deal with out-of-vocabulary words and to learn inflectional differences as well without additional modeling.

Poem translation has been tackled from the point of view of adaptation as well \cite{ghazvininejad2018neural}. The authors train a neural model to translate French poetry into English while making the output adapt to specified rhythm and rhyme patterns. They use an FSA (finite-state acceptor) to enforce a desired rhythm and rhyme.

Back-translation is also a viable starting point for style adaptation \cite{prabhumoye-etal-2018-style}. They propose a method consisting of two neural machine translation systems and style generators. They first translate the English input into French and then back again to English in the hopes of reducing the characteristics of the initial style. A style specific bi-LSTM model is then used to adapt the back translated sentence to a given style based on gender, political orientation and sentiment.

A recent line of work within the paradigm of computational creativity presents a creative contextual style adaptation in video game dialogs \cite{hamalainen2019creative}. They adapt video game dialog to better suit the state of the video game character. Their approach works in two steps: first, they use a machine translation model to paraphrase the syntax of the sentences in the dialog to increase the variety of the output. After this, they refill the new syntax with the words from the dialog and adapt some of the content words with a word embedding model to fit better the domain dictated by the player's condition.

A recent style adaptation \cite{li2019domain} learns to separate stylistic information from content information, so that it can maximize the preservation of the content while adapting the text to a new style. They propose an encoder-decoder architecture for solving this task and evaluate it on two tasks; sentiment transfer and formality transfer.

Earlier work on Finnish dialect normalization to standard Finnish has shown that the relationship between spoken Finnish varieties and literary standard language can be modeled as a character level machine translation task \cite{partanen2019dialect}.

\section{Data and Preprocessing}
\label{sec:data}

We use a corpus called Samples of Spoken Finnish \cite{skn} for dialect adaptation. This corpus consists of over 51,000 hand annotated sentences of dialectal Finnish. These sentences have been normalized on a word level to standard Finnish. This provides us with an ideal parallel data set consisting of dialectal text and their standard Finnish counterparts. 

The corpus was designed so that all main dialects and the transition varieties would be represented. The last dialect booklet in the series of 50 items was published in 2000, and the creation process was summarised there by \citeauthor{rekunen2000a} \shortcite{rekunen2000a}. For each location there is one hour of transcribed text from two different speakers. Almost all speakers are born in the 19th century. Transcriptions are done in semi-narrow transcription that captures well the dialect specific particularities, without being phonetically unnecessarily narrow.

The digitally available version of the corpus has a manual normalization for 684,977 tokens. The entire normalized corpus was used in our experiments.


\begin{table}[!h]
\centering
\begin{tabular}{|l|l|l|}
\hline
Dialect & Short & Sentences \\ \hline
Etelä-Häme & EH & 1860 \\ \hline
Etelä-Karjala & EK & 813 \\ \hline
Etelä-Pohjanmaa & EP & 2684 \\ \hline
Etelä-Satakunta & ES & 848 \\ \hline
Etelä-Savo & ESa & 1744 \\ \hline
Eteläinen Keski-Suomi & EKS & 2168 \\ \hline
Inkerinsuomalaismurteet & IS & 4035 \\ \hline
Kaakkois-Häme & KH & 8026 \\ \hline
Kainuu & K & 3995 \\ \hline
Keski-Karjala & KK & 1640 \\ \hline
Keski-Pohjanmaa & KP & 900 \\ \hline
Länsi-Satakunta & LS & 1288 \\ \hline
Länsi-Uusimaa & LU & 1171 \\ \hline
Länsipohja & LP & 1026 \\ \hline
Läntinen Keski-Suomi & LKS & 857 \\ \hline
Peräpohjola & P & 1913 \\ \hline
Pohjoinen Keski-Suomi & PKS & 733 \\ \hline
Pohjoinen Varsinais-Suomi & PVS & 3885 \\ \hline
Pohjois-Häme & PH & 859 \\ \hline
Pohjois-Karjala & PK & 4292 \\ \hline
Pohjois-Pohjanmaa & PP & 1801 \\ \hline
Pohjois-Satakunta & PS & 2371 \\ \hline
Pohjois-Savo & PSa & 2344 \\ \hline
\end{tabular}
\caption{Dialects and the number of sentences in each dialect in the corpus}
\label{tab:dialects-sentences}
\end{table}

Despite the attempts of the authors of the corpus to include all dialects, the dialects are not equally represented in the corpus. One reason for this is certainly the different sizes of the dialect areas, and the variation introduced by different speech rates of individual speakers. The difference in the number of sentences per dialect can be seen in Table \ref{tab:dialects-sentences}. We do not consider this uneven distribution to be a problem, as it is mainly a feature of this dataset, but we have paid attention to these differences in data splitting. In order to get proportionally even numbers of each dialect in the different data sets, we split the sentences of each dialect into training (70\%), validation (15\%) and testing (15\%) the split is done after shuffling the data. The same split is used throughout this paper.

The dialectal data contains non-standard annotations that are meant to capture phonetic and prosodic features that are usually not represented in the writing. These include the use of the acute accent to represent stress, superscripted characters, IPA characters and others. We go through all characters in the dialectal sentences that do not occur in the normalizations, i.e all characters that are not part of the Finnish alphabets and ordinary punctuation characters. We remove all annotations that mark prosodic features as these are not usually expressed in writing. This is done entirely manually as sometimes the annotations are additional characters that can be entirely removed and sometimes the annotations are added to vowels and consonants, in which case they form new Unicode characters and need to be replaced with their non-annotated counterparts.

\section{Automatic Dialect Adaptation}

\begin{table*}[!ht]
\centering
\begin{tabular}{|l|l|l|}
\hline
 & Source & Target \\ \hline
Flags & Inkerinsuomalaismurteet m i n ä \_ k u n \_ n ä i n & m i e \_ k o \_ n ä i n \\ \hline
No flags & m i n ä \_ k u n \_ n ä i n & m i e \_ k o \_ n ä i n \\ \hline
\end{tabular}
\caption{Example of the training data. The sentence reads "when I saw" in English}
\label{tab:training-data}
\end{table*}

In order to adapt text written in standard Finnish to dialects, we train several different models on the data set. As a character level sequence-to-sequence neural machine translation (NMT) approach has been proven successful in the past for the opposite problem of normalization of dialectal or historical language variant to the standard language (see \cite{bollmann-2019-large,hamalainen-etal-2019-revisiting,veliz2019benefits,hamalainen2019paft}), we approach the problem form a similar character based methodology. The advantage of character level models to word level models is their adaptability to out of vocabulary words; a requirement which needs to be satisfied for our experiments to be successful. In practice, this means splitting the words into characters separated by white-spaces and marking word boundaries with a special character, which is underscore (\_) in our approach.

In NMT, language flags have been used in the past to train multi-lingual models \cite{johnson-etal-2017-googles}. The idea is that the model can benefit from the information in multiple languages when predicting the translation for a particular language a expressed by a language specific flag given to the system. We train one model with all the dialect data, appending a dialect flag to the source side. The model will then learn to use the flag when adapting the standard Finnish text the the desired dialect.

Additionally, we train one model without any flags or dialectal cues. This model is trained to predict from standard Finnish to dialectal text (without any specification in terms of the dialect). This model serves two purposes, firstly if it performs poorly on individual dialects, it means that there is a considerable distance between each dialect so that a single model that adapts text to a generic dialect cannot sufficiently capture all of the dialects. Secondly, this model is used as a starting point for dialect specific transfer learning.

We use the generic model without flags for training dialect specific models. We do this by freezing the first layer of the encoder, as the encoder only sees standard Finnish, it does not require any further training. Then we train the dialect specific models from the generic model by continuing the training with only the training and validation data specific to a given dialect. We train each dialect specific model in the described transfer learning fashion for an additional 20,000 steps.

Our models are recurrent neural networks. The architecture consists of two encoding layers and two decoding layers and the general global attention model \cite{luong2015effective}. We train the models by using the OpenNMT Python package \cite{opennmt} with otherwise the default settings. The model with flags and the generic model are trained for 100,000 steps. We train the models by providing chunks of three words at a time as opposed to training one word or whole sentence at a time, as a chunk of three words has been suggested to be more effective in a character-level text normalization task \cite{partanen2019dialect}.

Table \ref{tab:training-data} shows an example of the sequences used for training. The model receiving the dialect flag has the name of the dialect appended to the beginning of the source data, where as the generic model has no additional information apart from the character sequences. The dialect specific transfer learning models are also trained without an additional flag, but rather the exposure solely to the dialect specific data is considered sufficient for the model to better learn the desired dialect.

\section{Results and Evaluation}

In this section, we present the results of the dialect adaptation models on different dialects. We use a commonly used metric called word error rate (WER) and compare the dialect adaptations of the test sets of each dialect to the gold standard. WER is calculated for each sentence by using the following formula:

\begin{equation}
WER = \frac{S + D + I}{S + D + C}
\end{equation}

WER is derived from Levenshtein edit distance \cite{Levenshtein66Binary} as a better measurement for calculating word-level errors. It takes into account the number of deletions $D$, substitutions $S$, insertions $I$ and the number of correct words $C$.

The results are shown in Tables \ref{tab:res1} and \ref{tab:res2}. On the vertical axis are the models. \textit{Flags} represents the results of the model that was trained with initial tokens indicating the desired dialect the text should be adapted in. \textit{No flags} is the model trained without any dialectal information, and the rest of the models are dialect specific transfer learning models trained on the \textit{no flags} model.

\begin{table*}[]
\centering
\tiny
\begin{tabular}{|l|l|l|l|l|l|l|l|l|l|l|l|l|}
\hline
model & EH & EK & EP & ES & ESa & EKS & IS & KH & K & KK & KP & LS \\ \hline
Flags & 24.37 & 19.8 & 25.13 & \textbf{28.09} & 27.22 & 25.19 & 21.09 & 28.73 & \textbf{25.56} & \textbf{24.59} & \textbf{22.51} & 30.49 \\ \hline
No flags & 38.87 & 36.21 & 41.98 & 42.16 & 37.71 & 37.35 & 39.38 & 39.03 & 37.05 & 42.43 & 39.08 & 42.3 \\ \hline
EH & \textbf{24.21} & 43.6 & 37.64 & 35.77 & 46.83 & 42.98 & 51.51 & 41.05 & 42.38 & 53.26 & 38.95 & 37.53 \\ \hline
EK & 48.65 & \textbf{19.28} & 52.63 & 47.57 & 35.69 & 39.94 & 31.86 & 42.97 & 47.14 & 33.13 & 49.76 & 45.51 \\ \hline
EP & 38.8 & 50.37 & \textbf{24.9} & 42.3 & 49.2 & 46.3 & 54.47 & 46.39 & 44.71 & 55.68 & 39.21 & 44.24 \\ \hline
ES & 34.36 & 44.81 & 41.49 & 29.03 & 49.35 & 47.8 & 50.05 & 45.56 & 47.74 & 51.16 & 38.02 & 37.12 \\ \hline
ESa & 46.06 & 32.28 & 49.5 & 50.38 & \textbf{26.81} & 32.43 & 42.01 & 44.26 & 38.4 & 40.32 & 45.88 & 47.9 \\ \hline
EKS & 44.3 & 37.3 & 47.06 & 51.05 & 34.15 & \textbf{25.07} & 45.56 & 42.97 & 36.5 & 42.84 & 42.65 & 47.86 \\ \hline
IS & 52.09 & 28.4 & 55.13 & 49.53 & 41.52 & 44.57 & \textbf{19.69} & 41.13 & 50.24 & 29.14 & 52.26 & 46.65 \\ \hline
KH & 43.98 & 38.34 & 47.75 & 47.66 & 45.46 & 43.23 & 41.16 & \textbf{28.43} & 47.88 & 44.36 & 47.9 & 45.76 \\ \hline
K & 42.59 & 45.05 & 45.11 & 50.11 & 39.79 & 35.97 & 50.56 & 48.17 & \textbf{25.56} & 49.34 & 40.89 & 49.63 \\ \hline
KK & 54.1 & 30 & 55.59 & 51.52 & 40.52 & 43.12 & 29.21 & 43.65 & 49.74 & 24.87 & 53.52 & 50.21 \\ \hline
KP & 35.58 & 43.94 & 38.58 & 40.2 & 44.54 & 41.53 & 51.03 & 45.84 & 39.26 & 52.04 & 22.51 & 44.32 \\ \hline
LS & 36.05 & 39.56 & 42.77 & 35.73 & 46.21 & 45.34 & 47.7 & 43.4 & 46.73 & 48.19 & 40.76 & \textbf{29.71} \\ \hline
LU & 38.45 & 45.07 & 44.24 & 39.17 & 51.68 & 51.03 & 47.35 & 41.04 & 51.14 & 49.54 & 46.74 & 38.97 \\ \hline
LP & 40.58 & 44.55 & 42.07 & 41.94 & 46.1 & 44.94 & 49.32 & 46.35 & 44.42 & 50.71 & 35 & 44.57 \\ \hline
LKS & 33.25 & 40.03 & 37.48 & 39.88 & 39.42 & 35.24 & 49.09 & 42.59 & 33.99 & 49.82 & 32.79 & 42.64 \\ \hline
P & 39.05 & 44.38 & 40.83 & 42.72 & 45.09 & 42.25 & 50.06 & 46.11 & 41.1 & 51.14 & 35.12 & 44.34 \\ \hline
PKS & 45.73 & 43.03 & 48.96 & 51.9 & 36.41 & 33.39 & 48.55 & 47.2 & 33.37 & 46.73 & 43.46 & 52.63 \\ \hline
PVS & 50.34 & 41.51 & 52.91 & 44.13 & 50.96 & 53.29 & 44.48 & 46.03 & 55.99 & 46.38 & 53.35 & 43.09 \\ \hline
PH & 31.26 & 44.72 & 38.38 & 37.56 & 44.61 & 39.4 & 52.07 & 42.19 & 38.51 & 52.73 & 35.43 & 40.82 \\ \hline
PK & 44.14 & 44.33 & 47.18 & 50.83 & 36.98 & 37.08 & 46.76 & 46.09 & 33.51 & 46.5 & 42.58 & 51.05 \\ \hline
PP & 34.73 & 44.38 & 37.87 & 41.85 & 43.24 & 39.46 & 52 & 45.12 & 36.91 & 52.84 & 27.12 & 43.25 \\ \hline
PS & 28.42 & 46.29 & 35.51 & 36.62 & 46.96 & 42.46 & 53.15 & 41.84 & 42.31 & 53.84 & 36.63 & 38.6 \\ \hline
PSa & 43.12 & 40.86 & 47.81 & 49.71 & 34.74 & 33.12 & 46.47 & 44.95 & 32.01 & 45.44 & 45.28 & 51.15 \\ \hline
\end{tabular}
\caption{WER for the different models for dialects from Etelä-Häme to Länsi-Satakunta}
\label{tab:res1}
\end{table*}

\begin{table*}[]
\centering
\tiny
\begin{tabular}{|l|l|l|l|l|l|l|l|l|l|l|l|}
\hline
model & LU & LP & LKS & P & PKS & PVS & PH & PK & PP & PS & PSa \\ \hline
Flags & 27.87 & 20.02 & 21.89 & 27.53 & 28.73 & 32.4 & \textbf{20.03} & \textbf{27.15} & 21.51 & \textbf{21.56} & 27.6 \\ \hline
No flags & 43.49 & 37.1 & 35.06 & 38.35 & 40.54 & 49.19 & 34.9 & 36.44 & 35.12 & 38.54 & 37.86 \\ \hline
EH & 39.9 & 35.63 & 33.65 & 39.92 & 48.42 & 51.05 & 27.61 & 41.9 & 32.54 & 27.46 & 43.91 \\ \hline
EK & 50.59 & 45.08 & 46.23 & 46.75 & 46.03 & 47.19 & 45.62 & 43.82 & 45.84 & 51.85 & 43.21 \\ \hline
EP & 47.04 & 37.78 & 39.13 & 41.56 & 52.06 & 56.16 & 33.28 & 44.64 & 34.32 & 35.23 & 46.35 \\ \hline
ES & 43.01 & 36.6 & 40.35 & 42.12 & 52.27 & 47.34 & 34.46 & 46.08 & 37.09 & 35.13 & 48.23 \\ \hline
ESa & 53.26 & 40.5 & 38.89 & 43.85 & 37.36 & 54.04 & 39.56 & 35.68 & 40.02 & 46.65 & 35.55 \\ \hline
EKS & 52.05 & 40.5 & 35.72 & 41.94 & 36.11 & 55.63 & 38.27 & 37.34 & 38.33 & 42.99 & 35.35 \\ \hline
IS & 48.72 & 47.29 & 49.67 & 48.39 & 49.59 & 45.74 & 49.89 & 46.45 & 48.51 & 54.29 & 46.18 \\ \hline
KH & 44.26 & 44.17 & 43.45 & 46.91 & 49.09 & 49.42 & 42.14 & 45.54 & 44.09 & 43.86 & 44.09 \\ \hline
K & 52.71 & 39.03 & 34.47 & 39.75 & 35.07 & 58.24 & 35.52 & 33.57 & 35.43 & 41.77 & 33.67 \\ \hline
KK & 51.83 & 48.19 & 50.94 & 49.37 & 49.41 & 48.7 & 50.84 & 45.66 & 50.27 & 55.14 & 45.86 \\ \hline
KP & 48.5 & 27.67 & 34.21 & 35.92 & 43.07 & 56.07 & 30.26 & 40.42 & 25.17 & 35.28 & 42.05 \\ \hline
LS & 42.57 & 36.9 & 39.88 & 42.58 & 51.71 & 47.31 & 34.74 & 46.31 & 38.71 & 36.28 & 47.7 \\ \hline
LU & \textbf{25.9} & 43.04 & 44.97 & 45.66 & 54.87 & 43.76 & 40.63 & 49.41 & 43.76 & 40.15 & 50.93 \\ \hline
LP & 49.41 & \textbf{19.57} & 38.2 & 32.23 & 47.75 & 55.87 & 35.04 & 43.57 & 33.96 & 37.92 & 45.85 \\ \hline
LKS & 45.39 & 33.2 & \textbf{21.41} & 34.97 & 40.88 & 55.13 & 26.9 & 36.47 & 28.22 & 31.57 & 36.37 \\ \hline
P & 47.29 & 28 & 35.54 & \textbf{26.81} & 46.23 & 56.06 & 33.45 & 40.98 & 32.7 & 37.82 & 43.27 \\ \hline
PKS & 55.67 & 41.86 & 39.87 & 42.62 & \textbf{28.65} & 57.68 & 39.28 & 35.8 & 38.08 & 46.01 & 33.67 \\ \hline
PVS & 46.42 & 49.26 & 54.99 & 52.31 & 57.36 & \textbf{31.67} & 52.13 & 52.69 & 51.34 & 53.44 & 53.39 \\ \hline
PH & 44.15 & 33.2 & 31.47 & 36.83 & 44.44 & 55.22 & \textbf{20.03} & 38.14 & 30.76 & 28.94 & 40.5 \\ \hline
PK & 53.77 & 41.01 & 38.61 & 42.59 & 38.34 & 57.18 & 37.99 & 27.28 & 37.87 & 46.02 & 34.98 \\ \hline
PP & 48.43 & 30.77 & 32.43 & 34.85 & 43.11 & 57.03 & 28.9 & 38.09 & \textbf{21.04} & 34.2 & 39.94 \\ \hline
PS & 42.17 & 35.18 & 32.03 & 38.9 & 47.14 & 54.2 & 26.49 & 41.34 & 31.54 & 22 & 42.98 \\ \hline
PSa & 52.19 & 42.13 & 36.28 & 42.45 & 35.29 & 56.52 & 38.29 & 32.8 & 37.67 & 43.96 & \textbf{27.24} \\ \hline
\end{tabular}
\caption{WER for the different models for dialects from Länsi-Uusimaa to Pohjois-Savo}
\label{tab:res2}
\end{table*}

The results are to be interpreted as the lower the better, i.e. the lower the WER, the closer the output is to the gold dialect data in a given dialect. These results indicate that the \textit{no flag} model does not get the best results for any of the dialects, which is to be expected, as if it reached to good results, that would indicate that the dialects do not differ from each other. Interestingly, we can observe that the transfer learning method gives the best scores for almost all the dialect, except for Etelä-Satakunta (ES), Keski-Karjala (KK), Keski-Pohjanmaa (KP), Pohjois-Karjala (PK) and Pohjois-Satakunta (PS), for which the model with flags gives the best results. Both methods are equally good for Pohjois-Häme (PH). All in all, the difference between the two methods is rather small in the WER. An example of the dialectal adaptation can be seen in Table \ref{tab:dial-examples}.

Based on these results it is difficult to suggest one method over the other as both of them are capable of reaching to the best results on different dialects. On the practical side, the model with dialectal flags trains faster and requires less computational resources, as the model is trained once only and it works for all the dialects immediately, where as transfer learning has to be done for each dialect individually after training a generic model.

Evaluation of the models with and
without dialectal flags shows that especially in word forms that are highly divergent in the dialect, it is almost impossible for the model to predict the correct result that is in the test set. This doesn't mean that the model's output would necessarily be entirely incorrect, as the result may still be perfectly valid dialectal representation, it just is in a different variety. 

There are also numerous examples of features that are in variation also within one dialect. In these cases the model may produce a form different from that in the specific row of a test set. These kind of problems are particularly prominent in examples where the dialectal transcription contains prosodic phenomena at the word boundary level. Since the model starts the prediction from standard Finnish input, it cannot have any knowledge about specific prosodic features of the individual examples in test data. Some phonological features such as assimilation of nasals seem to be over-generalized by the model, and also in this case it would be impossible for the model to predict the instances where such phenomena does not take place due to particularly careful pronunciation.

Another interesting feature of the model is that it seems to be able to generalize its predictions into unseen words, as long as they exhibit morphology common for the training data. There are, however, instances of clearly contemporary word types, such as recent international loans, that have general shape and phonotactics that are entirely absent from the training data. 
The problems caused by this are somewhat mitigated by fact that in many cases the standard Finnish word can be left intact, and it will pass within the dialectal text relatively well.

This has a consequence that the scores reported here are possibly slightly worse than the model's true abilities. The resulting dialectal text can still be very accurate and closely approximate the actual dialect, although the prediction would slightly differ from the test instances. At the same time if the model slips into predicted text some literary Finnish forms, the result is still perfectly understandable, and also in real use the dialects would rarely be used in entire isolation from the standard language. 

It must also be taken into account that only either a native dialect speaker or an advanced specialist in Finnish dialectology can reliably detect minute disfluencies in dialectal predictions, especially when the error is introduced by a form of other dialect. Similarly it would be very uncommon to have such knowledge about all the Finnish dialects the model operates on. After this careful examination of the models, we proceed to the generation of dialectal poems and their further evaluation by native Finnish speakers.

\begin{table*}[!h]
\scriptsize
\centering
\begin{tabular}{lllll}
\hline
SF & EK & PVS & IS & Translation \\ \hline
\begin{tabular}[c]{@{}l@{}}himo on palo,\\ se syttyy herkästi\\ taas intona se kokoaa\\ milloin into on eloisa?\\ näemmekö me,\\ ennen kuin into jää pois?\\ mikäli innot pysyisivät, \\ sinä huomaisit innon\\ minä alan maksamaan innon\\ olenko liiallinen?\end{tabular} & \begin{tabular}[c]{@{}l@{}}himo om palos,\\ se syttyy herkäst\\ taas intonna se kokovaa\\ millo into on elosa?\\ näämmekö met, \\ enne ku into jää pois?\\ mikäli innot pysysiit, \\ sie huomasit inno\\ mie alan maksamaa inno\\ olenko siialli?\end{tabular} & \begin{tabular}[c]{@{}l@{}}himo om palo,\\ se sytty herkästi\\ taas inton se kokko\\ millon innoo on elosa?\\ näämekö me, \\ ennen ku into jää pois?\\ mikäl innop pysysivät, \\ siä huamasit inno\\ mää ala maksaman inno\\ olenko liialline?\end{tabular} & \begin{tabular}[c]{@{}l@{}}himo om palloo,\\ se syttyy herkäst\\ toas inton se kokohoa\\ millon into on eloisa?\\ neämmäks meä, \\ ennen ku into jää pois?\\ mikält innot pysysiit, \\ sie huomaisit inno\\ mie ala maksamaa inno\\ olenko liialine?\end{tabular} & \begin{tabular}[c]{@{}l@{}}desire is a fire,\\ it gets easily ignited\\ again, as an ardor it shall rise\\ when is ardor vivacious?\\ will we see \\ before ardor disappears?\\ if ardors stayed, \\ you would notice the ardor\\ I will start paying for the ardor\\ Am I extravagant?\end{tabular} \\ \hline
\end{tabular}%
\caption{An example poem generated in standard Finnish and its dialectal adaptations to three different dialects}
\label{tab:dial-examples}
\end{table*}

\section{Effect on Perceived Creativity}

In this section, we apply the dialect adaptation trained in the earlier sections to text written in standard Finnish. We are interested in seeing what the effect of the automatically adapted dialect is on computer generated text. We use an existing Finnish poem generator \cite{hamalainen2018harnessing} that produces standard Finnish (SF) text as it relies heavily on hand defined syntactic structures that are filled with lemmatized words that are inflected with a normative Finnish morphological generator by using a tool called Syntax Maker \cite{hamalainen2018development}. We use this generator to generate 10 different poems.

The poems generated by the system are then adapted to dialects with the models we elaborated in this paper. As the number of different dialects is extensive and conducting human questionnaire with such a myriad of dialects is not feasible, we limit our study to three dialects. We pick Etelä-Karjala (EK) and Inkerinsuomalaismurteet (IS) dialects because they are the best performing ones in terms of WER and Pohjoinen Varsinais-Suomi (PVS) dialect as it is the worst performing in terms of WER. For this study, we use the dialect specific models tuned with transfer learning.

A qualitative look at the predictions revealed that the dialectal models have a tendency of over generating when a word chunk has less than three words. The models tend to predict one or two additional words in such cases, however, if the chunk contains three words, the models do not over nor under generate words. Fortunately this is easy to overcome by ensuring that only as many dialectal words are considered from the prediction as there were in the chunk written in standard Finnish. For instance \textit{olen vanha} (I am old) gets predicted in IS as \textit{olev vanha a}. The first two words are correctly adapted to the dialect, while the third word \textit{a} is an invention by the model. However, the models do not systematically predict too many words as in \textit{pieni ?} (small?) to \textit{pien ?} adaptation. For this reason, we only consider as many words as in the original chunks when doing the dialectal adaptation.

\subsection{Replicating the Poem Generator Evaluation}

In our first experiment, we replicate the poem generator evaluation that was used to evaluate the Finnish poem generator used in this experiment. We are interested in seeing whether dialectal adaptation has an effect on the evaluation results of the creative system. They evaluated their system based on the evaluation questions initially elaborated in a study on an earlier Finnish poem generator \cite{jukka}. The first evaluation question is a binary one \textit{Is the text a poem?}. The rest of the evaluation questions are asked on a 5-point Likert scale:

\begin{enumerate}
  \item How typical is the text as a poem?
  \item How understandable is it?
   \item How good is the language?
  \item Does the text evoke mental images?
  \item Does the text evoke emotions?
   \item How much do you like the text?
\end{enumerate}

The subjects are not told that they are to read poetry nor that they are reading fully computer generated and dialectally adapted text. We conduct dialectal adaptation to the 10 generated poems to the three different dialects, this means that there are altogether four variants of each poem, one in standard Finnish, and three in dialects. We produce the questionnaires automatically in such a fashion that each questionnaire has the 10 different poems shuffled in random order each time. The variants of each poem are picked randomly so that each questionnaire has randomly picked variant for each of the poems. Every questionnaire contains poems from all of the different variant types, but none of them contains the same poem more than once. Each questionnaire is unique in the order and combination of the variants. We introduce all this randomness to reduce constant bias that might otherwise be present if the poem variants were always presented in the same order.

We print out the questionnaires and recruit people native in Finnish in the university campus. We recruit 20 people to evaluate the questionnaires each of which consisting of 10 poems. This means that each variant of a poem is evaluated by five different people.

\begin{table*}[]
\centering
\tabcolsep=0.16cm
\small
\begin{tabular}{|l|c|c|c|c|c|c|c|c|c|c|c|c|c|c|c|c|c|c|c|}
\hline
 & Poem & \multicolumn{3}{c|}{Typical} & \multicolumn{3}{c|}{Understandable} & \multicolumn{3}{c|}{Language} & \multicolumn{3}{c|}{Mental images} & \multicolumn{3}{c|}{Emotions} & \multicolumn{3}{c|}{Liking} \\ \hline
 & \% & M & Mo & Me & M & Mo & Me & M & Mo & Me & M & Mo & Me & M & Mo & Me & M & Mo & Me \\ \hline
SF & 87.2\% & 2.85 & 4 & 3 & 3.62 & 4 & 4 & 3.51 & 4 & 4 & 3.57 & 4 & 4 & 2.94 & 2 & 3 & 3.02 & 4 & 3 \\ \hline
EK & 82.6\% & 2.5 & 2 & 2 & 3 & 4 & 3 & 2.87 & 3 & 3 & 3.26 & 4 & 3 & 2.67 & 2 & 2 & 2.70 & 2 & 3 \\ \hline
IS & 77.6\% & 2.69 & 2 & 3 & 2.90 & 3, 4 & 3 & 2.78 & 2 & 3 & 3.27 & 4 & 3 & 2.86 & 2 & 3 & 2.61 & 3 & 3 \\ \hline
PVS & 77.0\% & 2.51 & 2 & 2 & 2.80 & 2 & 3 & 2.58 & 2 & 3 & 3.27 & 4 & 3 & 2.86 & 2 & 3 & 2.61 & 3 & 3 \\ \hline
\end{tabular}
\caption{Results form the first human evaluation. Mean, mode and median are reported for the questions on Likert-scale}
\label{tab:humeval1}
\end{table*}

\begin{table}[!h]
\centering
\begin{tabular}{|l|l|l|l|}
\hline
 & EK & IS & PVS \\ \hline
WER & 34.38 & 43.41 & 54.69 \\ \hline
\end{tabular}
\caption{The distance of the dialectal poems form the original poem written in standard Finnish}
\label{tab:dist-original}
\end{table}

Table \ref{tab:humeval1} shows the results from this experiment, however some evaluators did not complete the task for all poems in their pile\footnote{The data is based on 47 observations for SF, 46 for EK, 43 for PVS and 49 for IS out of the maximum of 50.}. Interestingly, the results drop on all the parameters when the poems are adapted into the different dialects in question. The best performing dialect in the experiment was the Etelä-Karjala dialect, and the worst performing one was the Pohjoinen Varsinais-Suomi dialect all though it got the exact same average scores with Inkerinsuomalaismurteet on the last three questions. Now these results are not to be interpreted as that dialectal poems would always get worse results, as we only used a handful of dialects form the possibilities. However, the results indicate an interesting finding that something as superficial as a dialect can affect the results. It is to be noted that the dialectal adaptation only alters the words to be more dialectal, it does not substitute the words with new ones, nor does it alter their order.

In order to better understand why the dialects were ranked in this order, we compare the dialectal poems to the standard Finnish poems automatically by calculating WER. These WERs should not be understood as "error rates" since we are not comparing the dialects to a gold standard, but rather to the standard Finnish poems. The idea is that the higher the WER, the more they differ from the standard. Table \ref{tab:dist-original} shows the results of this experiment. The results seem to be in line with the human evaluation results; the further away the dialect is from the standard Finnish, the lower it scores in the human evaluation. This is a potential indication of familiarity bias; people tend to prefer the more familiar language variety.

\subsection{Word Association Test}

In the second experiment, we are interested in seeing how people associate words when they are presented with a standard Finnish version and a dialectally adapted variant of the same poem. The two poems are presented on the same page, labeled as A and B. The order is randomized again, which means that both the order of poems in the questionnarie and whether the dialectal one is A or B is randomized. This is done again to reduce bias in the results that might be caused by always maintaining the same order. The concepts we study are the following:

\begin{itemize}
    \item emotive
  \item original
  \item creative
   \item poem-like
  \item artificial
  \item fluent
\end{itemize}

The subjects are asked to associate each concept with A or B, one of which is the dialectal and the other the standard Finnish version of the same poem. We use the same dialects as before, but which dialect gets used is not controlled in this experiment. We divide each questionnaire of 10 poems into piles of two to reduce the work load on each annotator as each poem is presented in two different variant forms. This way, we recruit altogether 10 different people for this task, again native speakers from the university campus. Each poem with a dialectal variant gets annotated by five different people.

\begin{table}[!h]
\centering
\begin{tabular}{|l|l|l|l|}
\hline
 & SF & Dialect & No answer \\ \hline
emotive & \textbf{48\%} & 46\% & 6\% \\ \hline
original & 40\% & \textbf{60\%} & 0\% \\ \hline
creative & 32\% & \textbf{64\%} & 4\% \\ \hline
poem-like & 46\% & \textbf{50\%} & 4\% \\ \hline
artificial & \textbf{50\%} & 46\% & 4\% \\ \hline
fluent & \textbf{74\%} & 24\% & 2\% \\ \hline
\end{tabular}
\caption{Results of the second experiment with human annotators}
\label{tab:eval2-results}
\end{table}

Table \ref{tab:eval2-results} shows the results of this experiment. Some of the people did not answer to all questions for some poems. This is reflected in the \textit{no answer} column. The results indicate that the standard Finnish variant poems were considered considerably more fluent than the dialectal poems, and slightly more emotive and artificial. The dialectal poems were considered considerably more original and creative, and slightly more poem-like.

It is interesting that while dialectal poems can get clearly better results on some parameters on this experiment, they still scored lower on all the parameters in the first experiment. This potentially highlights a more general problem on evaluation in the field of computational creativity, as results are heavily dependent on the metric that happened to be chosen. The problems arising from this "ad hoc" evaluation practice are also discussed by \cite{lamb2018evaluating}.

\section{Conclusions}

We have presented our work on automatic dialect adaptation by using a character-level NMT approach. Based on our automatic evaluation, both the transfer learning method and a multi-dialectal model with flags can achieve the best results in different dialects. The transfer learning method, however, receives the highest scores on most of the dialects. Nevertheless, the difference in WERs of the two methods is generally small, therefore it is not possible to clearly recommend one over another to be used for different character-level data sets. If the decision is based on the computational power used, then the multi-dialectal model with flags should be used as it only needs to be trained once and it can handle all the dialects.

The dialect adaptation models elaborated in this paper have been made publicly available as an open-source Python library\footnote{https://github.com/mikahama/murre}. This not only makes the replication of the results easier but also makes it possible to apply these unique Finnish NLP tools on other related research or tasks outside of the academia as well. 

Our study shows that automatic dialect adaptation has a clear impact to how different attributes of the text are perceived. In the first experiment that was based on existing evaluation questions, a negative impact was found as the scores dropped on all the metrics in comparison to the original standard Finnish poem. However, when inspecting the distance the dialects have from the standard Finnish, we noticed that the further away the dialect is form the standard, the lower it scores. 

We believe that the low scores might be an indication of familiarity bias, which means that people have a tendency of preferring things they are more familiar with. Especially since the evaluation was conducted in a region in Finland with a high number of migrants from different parts of the country. This leads to a situation where the most familiar language variety for everyone regardless of their dialectal background is the standard Finnish variety. Also, as the dialectal data used in our model originates from the Finnish speakers born in the 19th century, it remains possible that the poems were transformed into a variety not entirely familiar to the individuals who participated into our survey. In the upcoming research it is necessary to investigate the perceptions of wider demographics, taking into account larger areal representation.


Based on our results, it is too early to generalize that familiarity bias is a problem in evaluation of computationally creative systems. However, it is an important aspect to take into consideration in the future research. We are interested in testing this particular bias out in the future in a more controlled fashion. Nevertheless, the fact that a variable, such as dialect that is never controlled in the computational creativity evaluations, has a clear effect on the evaluation results, raises a real question about the validity of such evaluation methods. As abstract questions on 5-point Likert scale are a commonly used evaluation methodology, the question of narrowing down the unexpected variables, such as dialect, that  affect the evaluation results positively or negatively is vital for the progress in the field in terms of comparability of results from different systems.

Even though the initial hypothesis we had on dialects increasing the perceived value of computationally created artefacts was proven wrong by the first experiment, the second experiment showed that dialects can indeed have a positive effect on the results as well, in terms of perceived creativity and originality. This finding is also troublesome form the point of view of computational creativity evaluation in a larger context. Our dialect adaptation system is by no means designed to exhibit any creative behavior of its own, yet people are more prone to associating the concept \textit{creativity} with dialectally adapted poetry.

The results of the first and second experiment give a very different picture of the impact dialect adaptation has on perceived creativity. This calls for a more thorough research on the effect different evaluation practices have on the results of a creative system. Is the difference in results fully attributable to subjectivity in the task, what was asked on how it was asked. Does making people pick between two (dialectal and standard Finnish in our case) introduce a bias not present when people rate the poems individually? It is important these questions be systematically addressed in the future research.






\section*{Acknowledgments}
Thierry Poibeau is partly supported by a PRAIRIE 3IA Institute fellowship
("Investissements d’avenir" program, reference ANR-19-P3IA-0001).

\bibliographystyle{iccc}
\bibliography{iccc}

\end{document}